\documentclass[letterpaper]{article} 
\usepackage{aaai2026}  
\usepackage{times}  
\usepackage{helvet}  
\usepackage{courier}  
\usepackage[hyphens]{url}  
\usepackage{graphicx} 
\usepackage{natbib}  
\usepackage{caption} 
\usepackage{algorithm}
\usepackage{amsmath,amsfonts,amssymb}
\usepackage{array}
\usepackage{textcomp}
\usepackage{verbatim}
\usepackage{xcolor}
\usepackage{diagbox}
\usepackage{caption}
\usepackage{enumitem}
\usepackage{pifont}
\usepackage{times}
\usepackage{soul}
\usepackage[utf8]{inputenc}
\usepackage{tikz}
\usepackage{bm}
\usepackage{algpseudocode}
\usepackage{booktabs}
\usepackage{siunitx}
\usepackage{xspace}
\usepackage{color}
\usepackage{colortbl}
\usepackage{multirow}
\usepackage{multicol}
\usepackage{subcaption}
\usepackage{cite}
\usepackage{placeins}

\usepackage{newfloat}
\usepackage{listings}
\DeclareCaptionStyle{ruled}{labelfont=normalfont,labelsep=colon,strut=off} 
\floatstyle{ruled}
\newfloat{listing}{tb}{lst}{}
\floatname{listing}{Listing}
\author{
    Qiyang Li\textsuperscript{\rm 1},
    Rui Kong\textsuperscript{\rm 1},%
    Yuchen Li\textsuperscript{\rm 1}\thanks{Corresponding Author},
    Hengyi Cai\textsuperscript{\rm 1},
    Shuaiqiang Wang\textsuperscript{\rm 1},
    Linghe Kong\textsuperscript{\rm 2},\\
    Guihai Chen\textsuperscript{\rm 2},
    Dawei Yin\textsuperscript{\rm 1}
}
\affiliations{
    \textsuperscript{\rm 1}Baidu Inc.\\
    \textsuperscript{\rm 2}Shanghai Jiao Tong University\\
}

\usepackage{xspace}

\newcommand{\sys}{AdaFuse\xspace}

\title{AdaFuse: Accelerating Dynamic Adapter Inference via Token-Level Pre-Gating and Fused Kernel Optimization}

\begin{document}

\maketitle
\begin{abstract}
The integration of dynamic, sparse structures like Mixture-of-Experts (MoE) with parameter-efficient adapters (e.g., LoRA) is a powerful technique for enhancing Large Language Models (LLMs). However, this architectural enhancement comes at a steep cost: despite minimal increases in computational load, the inference latency often skyrockets, leading to decoding speeds slowing by over 2.5 times. Through a fine-grained performance analysis, we pinpoint the primary bottleneck not in the computation itself, but in the severe overhead from fragmented, sequential CUDA kernel launches required for conventional dynamic routing.
To address this challenge, we introduce \sys, a framework built on a tight co-design between the algorithm and the underlying hardware system to enable efficient dynamic adapter execution. Departing from conventional layer-wise or block-wise routing, \sys employs a token-level pre-gating strategy, which makes a single, global routing decision for all adapter layers before a token is processed. This ``decide-once, apply-everywhere'' approach effectively staticizes the execution path for each token, creating an opportunity for holistic optimization. We capitalize on this by developing a custom CUDA kernel that performs a fused switching operation, merging the parameters of all selected LoRA adapters into the backbone model in a single, efficient pass. Experimental results on popular open-source LLMs show that \sys achieves accuracy on par with state-of-the-art dynamic adapters while drastically cutting decoding latency by a factor of over 2.4x, thereby bridging the gap between model capability and inference efficiency.
\end{abstract}
\section{Introduction}



Large language models (LLMs) have demonstrated remarkable capabilities in language understanding and generation, enabling significant progress in a wide range of tasks, including conversational AI~\cite{li2025towards,chen2025multi,li2023coltr,li2025fultr,wei2025igniting}, code generation~\cite{xiong2024search,wang2024enhancing}, search~\cite{li2025m,liao2023mcrle,li2025rankexpert,li2025s,li2025rankelectra}, and recommendation~\cite{li2023mhrr,li2023ltrgcn,lu2025dmmd4sr,cui2025multi,cui2025diffusion,liu2025mdn,mo2025one}. To customize the pretrained models to vertical domains or further enhance their capabilities, various adapter techniques such as Low-Rank Adapters (LoRA)~\cite{Hu2021LoRALA}, LLaMA-Adapter~\cite{zhang2023llama}, and Prompt Tuning~\cite{lester2021power,li2023s2phere,tong2025dapt,liao2024tpo} have been employed with great success. These methods are known for improving the performance of LLMs on downstream tasks without requiring extensive retraining, thereby enabling efficient model adaptation and customization.

Among these approaches, dynamic adapters~\cite{feng2024mixtureofloras,gou2024mixture,luo2024moelora} represent an even more potent strategy to augment the capacity of adapters. Unlike static adapters that apply the same transformation to all inputs, dynamic adapters conditionally activate different sets of lightweight modules based on input characteristics, enabling more sophisticated and context-aware model behavior.
By integrating conditionally computed lightweight adapters into the pretrained model, dynamic adapters allow for selective fine-tuning of adapter parameters.
This technique not only maintains the original strengths of the model but also substantially increases its adaptability and capacity across diverse tasks and domains. The flexibility to adapt different model components for different inputs theoretically provides superior performance compared to static approaches, making dynamic adapters particularly attractive for multi-task and multi-domain scenarios.

However, we found that despite the relatively minor impact of dynamic adapters on parameter size and computing complexity (typically adding only 1-5\% of the origin model), they may introduce significant latency overhead. 
For instance, the dynamic adapters that we studied all increase decoding inference latency by 250-950\%. 
The seemingly modest computational complexity of the low-rank matrices employed results in substantial extra CUDA kernel execution latency, surpassing that of models without dynamic adapters. 
This dramatic increase in latency is primarily attributed to the prolonged execution time of context operations during CUDA kernel runs, which considerably exceeds the actual computation time. 
The fundamental issue lies in the mismatch between the algorithmic design of dynamic adapters and the underlying GPU architecture, where frequent kernel launches and memory access patterns create substantial overhead that outweighs the computational benefits.
Dynamic adapters often require four or more additional CUDA kernel calls for each layer, in stark contrast to just a single call needed for the forward computation of the original backbone matrix. 
This excessive number of context operations substantially amplifies the latency overhead, leading to a severe escalation of inference latency.



Reducing the inference latency overhead of dynamic adapters is challenging. Existing dynamic adapters~\cite{dou2024loramoe,feng2024mixtureofloras,gao2024higher,gou2024mixture,li2024mixlora,liu2023moelora,luo2024moelora,wu2024parameterefficient,yang2024moral} adopt block-wise or layer-wise routing structures. This architectural choice inherently requires routing decisions to be made sequentially at each block or layer, preventing the efficient pre-merging of adapters into the backbone weights—a technique successfully used by static LoRA~\cite{Hu2021LoRALA}. The need to dynamically select and compute adapters at different stages of the model introduces a fundamental trade-off: the increased model expressiveness comes at the cost of fragmented, high-latency computations that are difficult to optimize with current system approaches. This makes it prohibitively costly to reduce the inference latency without altering the core design of dynamic routing.

\noindent\textbf{Our Approach: A System-Algorithm Co-Design.} Our approach to addressing the challenge is based on a holistic system-algorithm co-design. Specifically, we have developed a MoE-based dynamic adapters structure that facilitates token-wise adapter routing. Each token is associated with $k$ weighted paths of LoRA adapters, activated prior to the decoding of the token. 
This setup ensures that, although the model is enhanced with dynamic structures, the inference process for each token remains relatively static due to the pre-determined adapters.
To further enhance the efficiency, we pre-merge the activated LoRA adapters into the pretrained model's backbone before each token's decoding. 
This strategy fundamentally reduces the CUDA kernel execution overhead, thereby significantly lowering latency. 
With this innovative setup, we have re-engineered the inference process to seamlessly switch and merge adapters for each token, aligning the process closely with the original pretrained LLM's token decoding. 
Another pivotal component of our system is the development of a fused CUDA kernel, named SGMM, which efficiently manages the activated and inactivated adapters. 
This engineering solution ensures a smooth integration of dynamic adapters, optimizing both performance and efficiency.
%
%
We evaluate our \sys design across a range of benchmarks, comparing it against multiple state-of-the-art dynamic adapter baselines. The experiment results demonstrate that our approach are comparable with well-established strong baselines. Notably, our method significantly reduces the running overhead associated with other dynamic adapter alternatives, achieving an average speedup of 2.4 times in decoding latency.
In summary, our contributions are as follows:
\begin{itemize}
\item We uncover the high latency overhead introduced by dynamic adapters, which is a practical issue usually neglected by existing approaches. We analyze the fundamental reasons behind such high overhead, providing insights on the computational bottlenecks.
\item We introduce a novel architecture for dynamic adapters, named \sys. This design enhances the capacity of LLM adapters while minimizing the latency overhead, thereby offering an optimal balance between performance and efficiency.
\item Through extensive experiments, we demonstrate that \sys not only achieves accuracy on par with existing dynamic adapters across a variety of general and domain-specific tasks, but it also cuts down decoding inference latency by more than 2.4 times.
\end{itemize}


\section{Background and Motivation}


\begin{table*}[t]
    \centering
    
    \begin{tabular}{lccc}
    \toprule
    Method & Decoding latency (ms/token) & Parameter size (B) & FLOPS (G) \\
    \midrule
    Llama2-7B  & 2.4 & 6.74 & 6.61 \\
    MOLA~\cite{gao2024higher} & 25.3 (+954\%) & 7.07 (+4.89\%) & 6.65 (+0.61\%) \\
    PESC~\cite{wu2024parameter} & 8.5 (+254\%) & 6.97 (+3.41\%) & 6.64 (+0.45\%) \\
    MoRAL~\cite{yang2024moral} & 8.6 (+258\%) & 6.97 (+3.41\%) & 6.67 (+0.91\%) \\
    \bottomrule
    \end{tabular}
    
    \caption{Inference cost of different dynamic adapters.}
    \label{table:moa_latency}
\end{table*}

\noindent\textbf{Dynamic Adapters.}
Given the strengths of both the Mixture of Experts (MoE)~\cite{jiang2024mixtral,artic2024,dbrx2024,xAI-2024} and Low-Rank Adaptation (LoRA)~\cite{Hu2021LoRALA}, their integration has become a focal point of recent research efforts. Recent studies~\cite{feng2024mixtureofloras,gao2024higher,gou2024mixture,liu2023moelora,luo2024moelora} have explored combining these two techniques to further augment the capabilities of large language models (LLMs). This integration leverages the scalability of MoE and the efficiency of LoRA, proposing a promising pathway to meet the escalating demands for model performance and efficiency. 

Formally, the computation process of dynamic adapters can be formulated as:
\begin{align}
    y^l = f^l(x^l) + \sum_{i = 1}^{N}{G^l(x^l)_iE^l_i(x^l)},
    \label{eq:moa_origin}
\end{align}
where the superscript $l$ means $l$-th layer, $N$ represents number of adapters experts, $G^l(x^l) = \text{Softmax}(\text{TopK}(W^l_gx^l))$ represents the top-k (typically top-2) router in the dynamic adapters block, $f^l$ represents the pretrained backbone in $l$-th layer, and $E^l(x^l) = W^l_{up}(W^l_{down}(x^l))$ represents the output of LoRA experts.

Despite an increase in parameters, the experts of the dynamic adapters are activated sparsely, implying that only a limited subset of experts is used per input token. This sparse activation mechanism maintains computational efficiency while significantly expanding the model's capacity to handle diverse scenarios.

\noindent\textbf{Unexpected Latency Overhead of Dynamic Adapters.}
Although dynamic adapters can enhance accuracy and involve only a modest increase in parameter size and computing complexity, they unfortunately introduce a substantial inference latency overhead. We evaluate different dynamic adapter methods with Llama2-7B~\cite{touvron2023llama} on the ShareGPT~\cite{openchat2023sharegpt4} dataset for 50 queries one by one, and generate 200 new tokens for each query. As demonstrated in Table~\ref{table:moa_latency}, existing methods involving dynamic adapters result in an approximate 1\%-5\% increase in parameter count and less than a 1\% increase in computing complexity measured in FLOPS. However, these enhancements lead to a substantial increase in decoding latency, with overheads ranging from 200\% to 950\%.


\begin{figure}[t]  
    \centering
    \includegraphics[width=0.95\columnwidth]{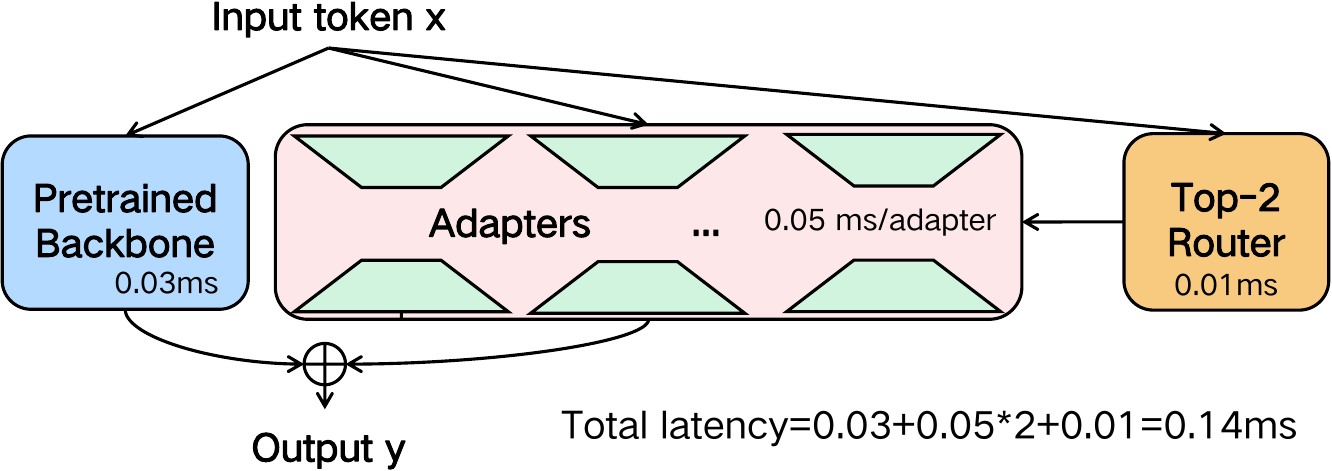}
    \caption{Decoding phase execution time profiling of one dynamic adapter layer in MoRAL. The execution time results were preceded by a warm-up of 100 executions and are obtained on the average of 300 executions.}
    \label{fig:cuda_launch}
\end{figure}

To elucidate the sources of latency overhead introduced by dynamic adapters, we conducted a granular analysis of latency within various components during the decoding phase. As illustrated in Figure~\ref{fig:cuda_launch}, it is evident that the execution time for the adapters (0.05ms) exceeds that of the pretrained backbone (0.03ms). Despite the relatively modest computational complexity of the LoRA adapters employed in dynamic configurations, each adapter necessitates dual launches of CUDA kernel context operations. The execution time of these CUDA kernels does not correlate linearly with the size of the matrices involved, leading to considerable latency in the adapter components.

\noindent\noindent\textbf{In-depth Latency Profiling.} To meticulously investigate how the inference latency of LoRA adapters and the pretrained backbone varies with increasing computational demands, we conducted tests across different input sequence lengths and examined the relationship between adapter rank and inference latency.

\begin{figure}[!h]
    \centering
    \begin{subfigure}[b]{0.45\textwidth}
        \centering
        \includegraphics[width=\textwidth]{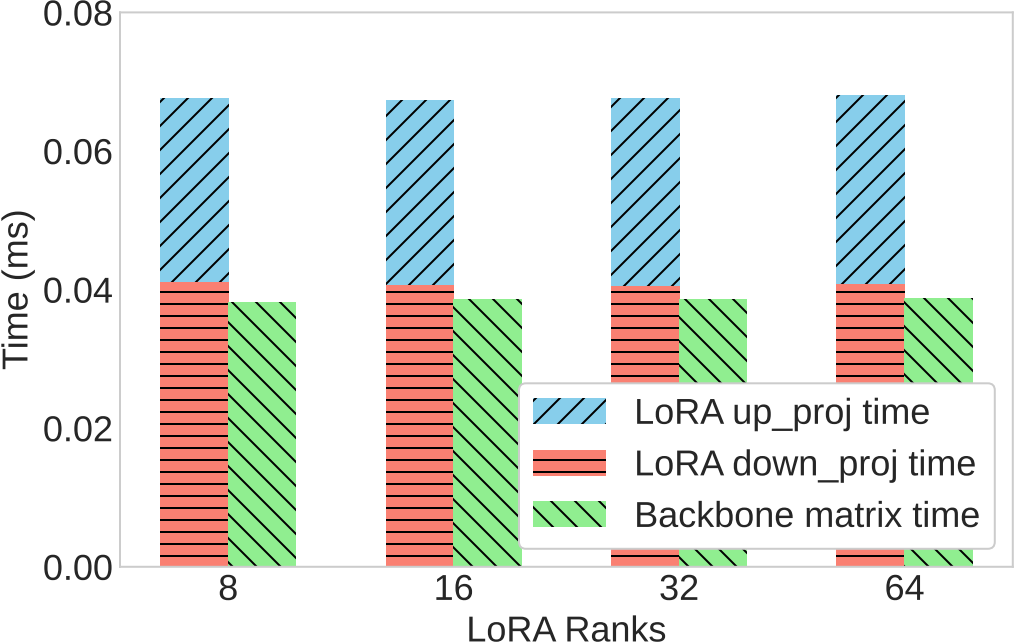}
        \caption{Prefilling phase: sequence length is 100.}
    \end{subfigure}
    \begin{subfigure}[b]{0.45\textwidth}
        \centering
    \includegraphics[width=\textwidth]{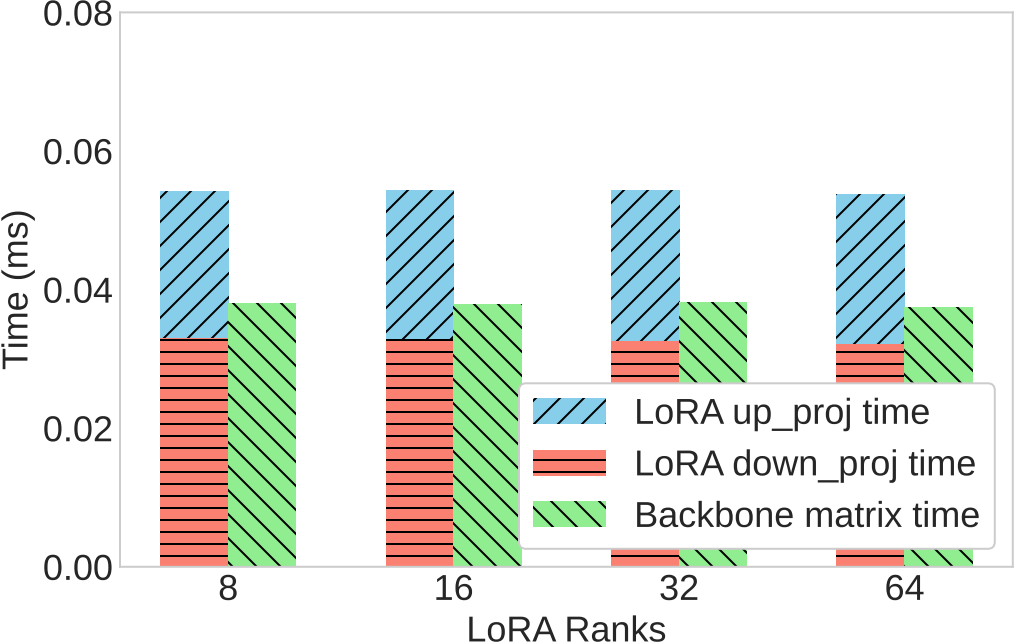}
        \caption{Decoding phase: sequence length is 1.}
    \end{subfigure}
    \caption{Latency breakdown of one dynamic adapter layer under different settings.}
    \label{figure:detailed adapter profiling}
\end{figure}

As shown in Figure~\ref{figure:detailed adapter profiling}, regardless of whether it is during the prefilling or decoding phase, and irrespective of the LoRA ranks being high or low, the latency of the LoRA adapters consistently exceeds that of the backbone. This phenomenon is primarily attributed to the number of CUDA kernel calls rather than the computing complexity involved in each call. The underlying reason is that the latency associated with CUDA kernel calls does not scale linearly with computing complexity. This insight highlights a crucial aspect of system behavior that significantly impacts the performance of dynamic adapters.

\noindent\noindent\textbf{Challenge of Reducing Latency Overhead.}
A straightforward way to reduce inference latency overhead is to reduce the times of CUDA kernel context operations. Like LoRA~\cite{Hu2021LoRALA}, one could pre-merge adapters into the original matrix and then perform token decoding computation. We use this simple strategy in MoRAL by directly merging activated adapters layer by layer before computing. However, the additional operations introduced higher latency, where the decoding latency is 4.5 ms/token, which is still 88\% higher than the original LLM model. This is because merging a LoRA adapter into the backbone matrix requires an additional invocation of a CUDA kernel to perform the matrix multiplication for the up and down projections.

\section{Related Work}
\noindent\textbf{Parameter Efficient Fine-tuning (PEFT).}
Parameter-efficient fine-tuning has emerged as the predominant approach for adapting pretrained large language models (LLMs) to downstream tasks. Representative methods include Adapters~\cite{houlsby2019parameter}, Prefix Tuning~\cite{Li2021PrefixTuningOC}, Prompt Tuning~\cite{Lester2021ThePO}, and LoRA~\cite{Hu2021LoRALA}. Among these, LoRA has gained prominence due to its elegant low-rank decomposition strategy that enables efficient inference through adapter merging. However, traditional LoRA is limited to static single-adapter scenarios, lacking the dynamicity needed for multi-task adaptation.

\noindent\textbf{Dynamic Adapters and MoE-based PEFT.}
The convergence of PEFT and MoE principles has led to dynamic adapter architectures. Block-wise methods such as MOLA~\cite{gao2024higher} and MoELoRA~\cite{luo2024moelora} require runtime routing within each block, preventing efficient batching. Layer-wise approaches like MoRAL~\cite{yang2024moral} and LoRAMoE~\cite{dou2024loramoe} apply routing at layer granularity but fundamentally cannot achieve the inference efficiency of static adapters due to their layer-by-layer routing decisions. In contrast, our token-wise pre-gated approach makes all routing decisions upfront, enabling adapter merging before inference and eliminating computational overhead entirely.

\noindent\textbf{System Optimizations for LLM Inference.}
System-level optimizations for LLMs include memory management techniques like PagedAttention~\cite{kwon2023efficient}, quantization~\cite{Frantar2022GPTQAP}, and pruning~\cite{Frantar2023SparseGPTML}. PEFT-specific optimization ~\cite{Ye2023ASPENHL} targets static adapter configurations but cannot address dynamic adapter selection due to irregular computation patterns. Our work uniquely addresses the gap between algorithmic innovations in dynamic adapters and system-level optimizations by redesigning the dynamic adapter paradigm to be system-friendly while preserving expressiveness.


\section{Design of \sys}

\begin{figure*}[!h]
    \centering
    \includegraphics[width=1\linewidth]{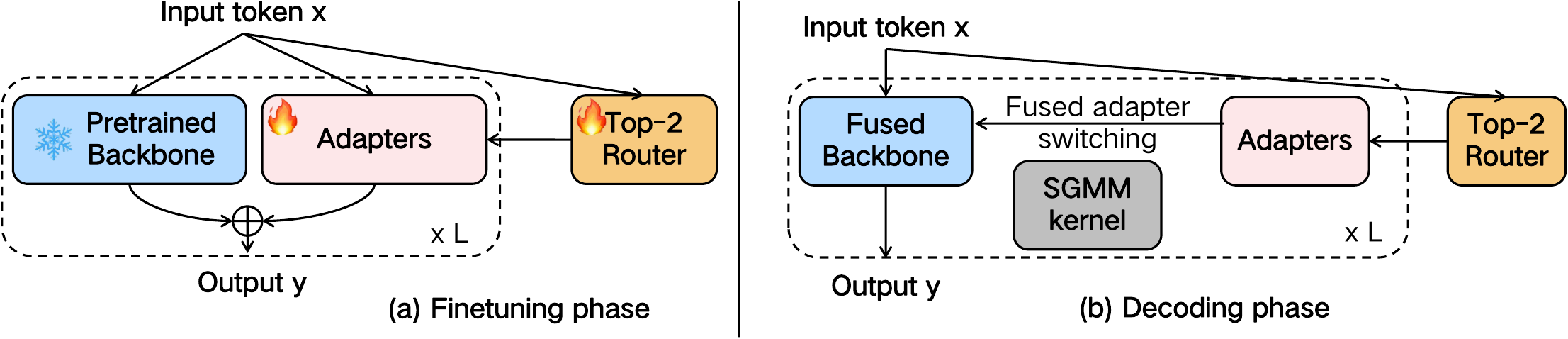}
    \caption{Overview of \sys.}
    \label{fig:overview}
\end{figure*}

\subsection{Overview}

As shown in Figure~\ref{fig:overview}, given a pre-trained LLM, we first extend it with \sys to enhance its model capacity and then finetune it on either general or domain-specific datasets. 
The core innovation of \sys lies in its token-wise pre-gated architecture, which fundamentally differs from existing layer-wise or block-wise dynamic adapter approaches. Instead of making routing decisions at each layer independently, \sys employs a single router at the first layer to determine expert activation patterns for all subsequent layers, thereby eliminating the computational overhead of repeated routing computations.

During the decoding phase, each token is initially processed through a router to compute the gating for each layer. The key insight is that once we determine which experts should be activated for a given token, this decision can be propagated across all layers without requiring additional routing computations. This design choice is motivated by our observation that tokens with similar semantic properties tend to activate consistent expert patterns across different layers.
To implement this efficiently, we developed an SGMM kernel that facilitates fast fused adapter switching. This advanced functionality enables the rapid merging of activated adapters into the backbone and the efficient unmerging of inactivated adapters from the backbone, significantly reducing the number of CUDA kernel calls.

\subsection{Model Structure} \label{section:method:structure}

In \sys, we extend adapters only in the linear layers of the pre-trained backbone, and we insert a Top-2 router $G^1$ only at the first expanded linear layer. This architectural choice is carefully designed to balance computational efficiency and model expressiveness.

\noindent\textbf{Router Architecture.} The Top-2 router $G^1$ consists of a lightweight linear transformation followed by a softmax activation and top-k selection mechanism. The router takes the hidden representation from the first layer as input and produces routing weights that determine expert activation across all layers.

\noindent\textbf{Expert Configuration.} Each expert in \sys is implemented as a LoRA module with rank $r$ typically set to 64 or 128, depending on the model size and task complexity. The number of experts $N$ is configurable and typically ranges from 4 to 16, providing a good balance between model capacity and computational overhead.

\noindent\textbf{Finetuning Phase.}
As shown in Figure~\ref{fig:overview} (a), during fine-tuning, the computation process can be written as:
\begin{equation}
    y^l = f^l(x^l) + \sum_{i = 1}^{N}{G^1(x^{1})_iE^l_i(x^l)}
    \label{equation:ours:definition}
\end{equation}
where \sys only replaces the $G^l(x^l)$ with $G^1(x^{1})$ in Equation~\ref{eq:moa_origin}, where $x^{1}$ denotes the input in the first expanded linear layer. \sys employs a token-wise pre-gated LoRA structure, meaning that the routing weights for all layer adapters are identical. 
This design not only preserves the model's dynamicity but also facilitates latency optimization during inference.

\noindent\textbf{Decoding Phase.}
As depicted in Figure~\ref{fig:overview} (b), \sys leverages the token-wise pre-gated structure to reduce latency during the decoding phase. During the decoding phase of LLM generation, where the input \( x \) is a single token, the Top-2 router $G^1$ determines which experts' adapters are activated. 
Then the activated experts are merged into pretrained backbone. Finally, the fused backbone performs forward process and the decoding computation as:
\begin{align}\label{equation:ours:1}
    y^l = f^l_*(x^l)
\end{align}
To efficiently calculate $f^l_*$ for all layers, we propose to perform fused adapter switching with the SGMM kernel, which merges the parameters of all activated expert adapters across all layers into the original parameters of the pretrained model in a single CUDA kernel operation. 
Finally, the fused backbone execute just like the initial pretrained backbone as shown in Equation~\ref{equation:ours:1}.

\noindent\textbf{Prefilling Phase.}
The latency of \sys during the prefilling phase is comparable to that of existing dynamic adapters. For the prefilling phase, we have not implemented specific optimizations, as the latency in the LLM generation stage primarily originates from the decoding phase.

\subsection{Fused Adapter Switching}    \label{section:method:switching}
To calculate $f^l_* = f^l + \sum_{i = 1}^{N}{G^1(x^1)_i} E^l_i$ according to Equation~\ref{eq:moa_origin} and Equation~\ref{equation:ours:definition}, we may merge multiple experts adapters into backbone because one input token may activate multiple adapters (typically Top-2). The experts in \sys are LoRA adapters, which contain down projection \text{LoRA\_DOWN} and up projection \text{LoRA\_UP}.

We only need to invoke the CUDA kernel $k$ times instead of $N$ times, as $G^1(x^1)$ specifically targets the top-$k$ selections. To further reduce the number of CUDA kernel calls, we concatenate all LoRA adapters before merging them into the original model parameters as:
\begin{flalign}\label{equation:ours:lora_unified_single}
    &\text{LoRA\_DOWN}^l = \text{concat}\big[G^1(x^1)_i \cdot \text{LoRA\_DOWN}^l_i, \nonumber \\
    &\hfill i=1,\ldots,N\big] & \\
    &\text{LoRA\_UP}^l = \text{concat}\big[\text{LoRA\_UP}^l_i, i=1,\ldots,N\big] &
\end{flalign}

Thus, we can merge the concatenated LoRA adapters into origin backbone as:
\begin{align}\label{equation:ours:lora merge2}
    f^l_* = f^l + \text{LoRA\_DOWN}^l \times \text{LoRA\_UP}^l
\end{align}

We observe that during the decoding phase, the activated adapters varying from one token to the next. A simple way to obtain $f^l$ is to unmerge the activated adapters by the last token in the current iteration. Then we concatenate the activated adapters of current input and inactivated adapters of last input as:
\begin{align}\label{equation:ours:lora_unified_fused}
    &\text{Fused\_LoRA\_DOWN}^l = \text{concat}\big[-(\text{LoRA\_DOWN}^l)^{t-1}, \nonumber \\
    &\hfill (\text{LoRA\_DOWN}^l)^{t}\big] \\
     &\text{Fused\_LoRA\_DOWN}^l = \text{concat}\big[-(\text{LoRA\_DOWN}^l)^{t-1}, \nonumber \\
    &\hfill (\text{LoRA\_DOWN}^l)^{t}\big]   
\end{align}

So the concatenated LoRA adapter switching operation can be rewritten as:

\begin{equation}\label{equation:ours:fused}
\begin{split}
    (f^l_*)^t &= (f^l_*)^{t-1} + \quad \text{Fused\_LoRA\_DOWN}^l \times \\ &\text{Fused\_LoRA\_UP}^l
\end{split}
\end{equation}

To calculate Equation~\ref{equation:ours:fused}, we utilize our efficiently designed CUDA kernel, SGMM, to seamlessly integrate these fused adapters into the pretrained LLM backbone for all layers with only one CUDA kernel call.

\subsection{SGMM Kernel}    \label{section:method:sgmm}

The straight-forward way to merge the LoRA adapter into the backbone is to merge them layer by layer. This approach requires multiple calls to the CUDA kernel, which introduces additional latency due to kernel launches. Moreover, smaller kernel computations underutilize GPU thread blocks, leading to a low GPU throughput. We observe the layer-by-layer merging operations can be handled concurrently and introduce a CUDA kernel called Segmented Gather Matrix Multiplication (SGMM) to finally handle the merging of LoRA adapters of \sys, adapted from the concept of SGMV proposed by Punica \cite{chen2023punica}.

SGMM is designed to execute a batched GEMM operations, which can be summarized by the following equation:
\begin{equation}\label{equation:sgmm}
    f_* =  f +  \text{Fused\_LoRA\_DOWN} \times \text{Fused\_LoRA\_UP},
\end{equation}
where $f_*$ is the resultant updated matrix of the backbone; $f$ is the original weight matrix of the backbone; $\text{Fused\_LoRA\_DOWN}$ and $\text{Fused\_LoRA\_UP}$ are the adapter matrices for weight matrix $f$. The addition operation within the SGMM kernel is performed in place, significantly reducing the additional memory overhead.

When wrapping these operations into a single CUDA kernel, taking full advantage of the GPU's computational resources is challenging. To achieve this, we divide the matrix multiplication into multiple GEMM tiles and assign them to different thread blocks. On the other hand, these thread blocks must switch context with global memory/shared memory frequently, thus causing significant latency. To tackle this, we adopt a pre-fetch buffer mechanism to hide loading latency.

The input parameters for SGMM are arrays of pointers to the LoRA matrices and the backbone matrices, which respectively store the corresponding entries of each layer's LoRA matrix and the shape of each matrix segment. When launching the kernel, SGMM applies as many thread blocks as possible and divides the large matrix multiplication into multiple GEMM tiles of the same shape, with each tile operating matrix computation. The optimal tiling scheme related to hardware is selected to ensure the full utilization of each thread block. This parallel execution enables the efficient merging and unmerging of LoRA weights, a critical operation in our approach.


\section{Evaluation}

\renewcommand{\topfraction}{0.9}
\renewcommand{\bottomfraction}{0.8}
\setcounter{topnumber}{2}
\setcounter{bottomnumber}{2}
\setcounter{totalnumber}{4}
\renewcommand{\dbltopfraction}{0.9}
\setcounter{dbltopnumber}{2}


\subsection{Experiment Setup}

\noindent\textbf{Datasets/benchmarks.}
To demonstrate the proposed \sys could augment general ability of LLM, we follow PESC~\cite{wu2024parameter} and simultaneously fine-tuned the model on a diverse set of skills, including encompassing coding, mathematical, and other general abilities from various subjects. This training involved integrating three distinct datasets from varied domains during the instruction tuning phase: SlimORCA~\cite{SlimOrca},  Magicoder~\cite{wei2023magicoder}, and MetaMathQA~\cite{yu2023metamath} datasets.
We utilize LM-Eval-Harness~\cite{eval-harness} as tool to evaluate general ability on ARC~\cite{Clark2018ThinkYH}, HellaSwag~\cite{zellers2019hellaswag}, MMLU~\cite{hendryckstest2021}, TruthfulQA~\cite{lin-etal-2022-truthfulqa}, WinoGrande~\cite{sakaguchi2019winogrande}, and MT-Bench~\cite{zheng2023judging} benchmarks and report the accuracy.
Also, to demonstrate the proposed \sys could improve domain specific ability of LLM, we follow MoLA~\cite{gao2024higher} and fine-tuned the model on downstream task. We evaluate three recent question-answering benchmarks, including ScienceQA\cite{lu2022learn}, CommonsenseQA\cite{talmor-etal-2019-commonsenseqa}, and OpenbookQA\cite{OpenBookQA2018}.
To evaluate runtime efficiency performance of \sys, we utilize real-world sharegpt~\cite{openchat2023sharegpt4} dataset to simulate user queries. We serve 50 queries from sharegpt dataset one by one, and generate 200 new tokens for each query.

\noindent\textbf{Baselines.}
We compare \sys with four PEFT approaches, including LoRA~\cite{Hu2021LoRALA}, layer-wise gating dynamic adapters like MoRAL~\cite{yang2024moral} and MOLA~\cite{gao2024higher}, and block-wise gating dynamic adapters PESC~\cite{wu2024parameter}. We use LLama2-7B~\cite{Touvron2023Llama2O} and Mistral-7B~\cite{jiang2023mistral} as the pretrained base LLM.

\subsection{Accuracy Evaluation}

We evaluate the accuracy of \sys on both general capabilities and domain-specific tasks to demonstrate its effectiveness compared to existing dynamic adapter methods.

\noindent\textbf{General Capability Enhancement.}
To assess the general capability improvement, we fine-tune \sys and baseline methods on a mixture of general datasets and evaluate on standard benchmarks. Table~\ref{table:general accuracy} shows the results across five common evaluation tasks.

\begin{table}[htbp] 
    \centering
    
    {
        \scriptsize 
        \setlength{\tabcolsep}{2pt} 
        
        \begin{tabular}{lccccccc}
        \toprule
        Method & ARC & HellaSwag & MMLU & TruthfulQA & Winogrande & Avg  \\
        \midrule
        Llama2-7B (base) & 51.71 & \textbf{77.74} & 48.30 & 45.31 & 72.45 & 59.10 \\
        LoRA & 51.79 & 77.02 & 50.46 & 45.13 & 73.80 & 59.64 \\
        MoRAL (layer-wise) & 52.13 & 77.57 & 51.10 & 45.93 & \textbf{74.35} & 60.22 \\
        PESC (block-wise) & \textbf{53.58} & 77.27 & 51.07 & 46.04 & 74.27 & \textbf{60.45} \\
        \sys (ours) & 52.39 & 77.60 & \textbf{51.15} & \textbf{46.15} & 73.32 & 60.12 \\
        \bottomrule
        \end{tabular}
    } 
    
    \caption{Accuracy of incremental training achieved with different dynamic adapters.}
    \label{table:general accuracy}
\end{table}

As shown in Table~\ref{table:general accuracy}, \sys achieves competitive performance with an average accuracy of 60.12\%, which is comparable to the best performing baseline PESC (60.45\%). Notably, \sys achieves the highest scores on MMLU and TruthfulQA benchmarks, demonstrating its effectiveness in knowledge-intensive tasks.

\noindent\textbf{Domain-Specific Customization.}
We further evaluate \sys on domain-specific tasks to assess its adaptation capability. Tables~\ref{table:specific accuracy} and~\ref{table:specific accuracy2} present results on question-answering tasks for Llama2-7B and Mistral-7B respectively.

\begin{table}[htbp] 
    \centering
    
    {
        \scriptsize
        \setlength{\tabcolsep}{2pt}
        
        \begin{tabular}{lcccc}
        \toprule
        Methods & ScienceQA & CommonsenseQA & OpenbookQA & Avg \\
        \midrule
        Llama2-7B (base) & 53.19 & 47.82 & 45.80 & 48.94 \\
        Full-Parameter & 93.12 & 77.48 & 80.40 & 83.67 \\
        LoRA & 91.01 & 75.51 & 77.00 & 81.17 \\
        MoLA (layer-wise) & \textbf{91.91} & 77.89 & \textbf{82.80} & \textbf{84.20} \\
        MoRAL (layer-wise) & 90.74 & 76.41 & 76.60 & 81.25 \\
        PESC (block-wise) & 90.02 & 76.00 & 78.40 & 81.47 \\
        \sys (ours) & 91.39 & \textbf{79.03} & 80.40 & 83.60 \\
        \bottomrule
        \end{tabular}
    } 
    
    \caption{Accuracy of domain-specific fine-tuning achieved with different dynamic adapters on Llama2-7B.}
    \label{table:specific accuracy}
\end{table}

\begin{table}[htbp]
    \centering
    
    {
        \scriptsize
        \setlength{\tabcolsep}{2pt}
        
        \begin{tabular}{lcccc}
        \toprule
        Methods           & ScienceQA & CommonsenseQA & OpenbookQA & Avg \\
        \midrule
        Mistral-7B (base) & 62.24             & 58.93                 & 57.8               & 59.66        \\
        LoRA              & 94.15             & 79.85                 & 84.2               & 86.06        \\
        MoRAL (layer-wise) & 93.79             & 81.57                 & 85.8               & 87.05        \\
        PESC (block-wise) & 94.33             & 80.46                 & 86.4               & 87.06        \\
        \sys (ours)       & 93.82             & 81.29                 & 86.6               & \textbf{87.24} \\
        \bottomrule
        \end{tabular}
    } 
    
    \caption{Accuracy of domain-specific fine-tuning achieved with different dynamic adapters on Mistral-7B.}
    \label{table:specific accuracy2}
\end{table}

For domain-specific tasks, \sys demonstrates strong performance, achieving 83.60\% average accuracy on Llama2-7B and 87.24\% on Mistral-7B, both competitive with or exceeding baseline methods. Particularly noteworthy is \sys's superior performance on CommonsenseQA, suggesting effective reasoning capability enhancement.

\subsection{Runtime Performance}

Beyond accuracy, a critical evaluation metric for dynamic adapters is their runtime efficiency. We conduct comprehensive latency and memory usage analysis to demonstrate the practical advantages of \sys over existing methods.


\begin{table}[t] 
    \centering
    \footnotesize 
    
    \setlength{\tabcolsep}{3.5pt} 
    
    \begin{tabular}{lcc}
    \toprule
    Method & Latency (ms/token) & Memory (GiB) \\
    \midrule
    Llama2-7B & 2.4 & 12.9\\
    MOLA (layer-wise) & 25.3 (+954\%) & 26.3 (+104\%)\\
    PESC (block-wise) & 8.5 (+254\%) & 13.1 (+2\%)\\
    MoRAL (layer-wise) & 8.6 (+258\%) & 13.3 (+3\%)  \\
    \sys (ours) & 3.1 (+29\%) & 13.8 (+7\%)\\
    \bottomrule
    \end{tabular}
    
    \caption{Decoding latency and peak memory overhead of different dynamic adapters.}
    \label{table:latency}
\end{table}

As reported in Table~\ref{table:latency}, we evaluated the inference latency and peak GPU memory usage of our method and various baseline methods on the ShareGPT dataset. The results show that \sys exhibits significantly lower decoding latency compared to all other dynamic adapter methods, being 2.7 times faster than the previously fastest method, PESC. Furthermore, \sys's decoding latency is less than 30\% higher than that of the original Llama2-7B model.
The dramatic latency reduction achieved by \sys can be attributed to our fundamental architectural innovation. While traditional dynamic adapters require separate routing computations and adapter activations at each layer, our approach consolidates these operations into a single pre-gating decision followed by efficient batch processing through the SGMM kernel. This transforms the inference pattern from multiple sequential operations to a single parallelized computation.

\subsection{Ablation Study}

To understand the contribution of individual components in \sys, we conduct ablation studies focusing on the key system-level optimization: the SGMM kernel. 

\begin{table}[htbp]
    \centering
    \begin{tabular}{lc}
    \toprule
    Method & Latency (ms/token) \\
    \midrule
    Llama2-7B & 2.4 \\
    MoRAL & 8.5 (+254\%) \\
    MoRAL (Simple merge) & 4.5 (+88\%) \\
    \sys (Simple merge) & 4.2 (+ 75\%) \\
    \sys  & 3.1 (+29\%) \\
    \bottomrule
    \end{tabular}
    
    \caption{Ablation study for runtime decoding latency of \sys.}
    \label{table:ablation study latency}
\end{table}

Table~\ref{table:ablation study latency} demonstrates the critical importance of the SGMM kernel in achieving our performance gains. Replacing the SGMM with a simple merge approach results in a substantial increase in decoding latency for \sys, rising from 3.1 ms/token to 4.2 ms/token. This validates our hypothesis that kernel-level optimizations are essential for dynamic adapter efficiency.

The simple merge approach, while algorithmically equivalent, fails to exploit the parallelism opportunities inherent in our fused adapter switching strategy. We also tested this simple merge technique on MoRAL, which reduced its latency to 4.5 ms/token but still registered an 88\% increase over the original Llama2-7B. These findings highlight the effectiveness of our system-algorithm co-design approach.


\FloatBarrier

\section{Conclusion}

This paper addresses the critical inference latency overhead (250-950\%) in dynamic LLM adapters, which we identify as a consequence of fragmented CUDA kernel calls, not computational load. We propose \sys, a system-algorithm co-design featuring a token-wise pre-gated architecture. \sys uses a single router at the first layer to determine expert activations for all subsequent layers, a strategy that eliminates the overhead of traditional layer-wise routing.
To realize this design, we developed the SGMM CUDA kernel to perform fused adapter switching, which merges activated adapters into the backbone in a single efficient operation. Rigorous experiments validate our approach: \sys achieves competitive accuracy (60.12\% general, 83.58\% domain-specific) while delivering a 2.4× decoding speedup over the fastest dynamic adapters. This reduces the latency overhead from a typical 250-950\% to just 29\% over the backbone model.
Ablation studies confirm that both the pre-gated architecture and our SGMM kernel are essential to this performance. Ultimately, \sys establishes a new benchmark for efficient LLM tuning, providing a practical solution that maintains model adaptability while achieving near-native inference speeds and contributing new design principles for future dynamic adapter architectures.

\bibliography{aaai2026}

\end{document}